\documentclass{article}
\usepackage{ijcai21}

\usepackage{soul}
\usepackage{url}
\usepackage[hidelinks]{hyperref}
\usepackage[utf8]{inputenc}
\usepackage[small]{caption}
\usepackage{graphicx}
\usepackage{amsmath}
\usepackage{amsthm}
\usepackage{booktabs}
\usepackage{algorithm}
\usepackage{algorithmic}

\usepackage{subcaption}
\usepackage{booktabs}
\usepackage{url}
\usepackage{todonotes}
\usepackage{multirow}
\usepackage{tikz}

\usepackage{amsfonts}
\usepackage{thmtools}
\usepackage{mleftright}

\theoremstyle{theorem}

\newtheorem{insight}{Insight}

\usepackage{thm-restate}
\usepackage[mathic=true]{mathtools}
\usepackage{fixmath}

\usepackage{pifont}

\usepackage{enumitem}
\setlist[enumerate]{noitemsep, topsep=0.5\topsep}
\setlist[description]{noitemsep, topsep=0.5\topsep}
\setlist[itemize]{noitemsep, topsep=0.5\topsep}

\usepackage{setspace}
\usepackage[protrusion=true,expansion=false, activate={true,nocompatibility},final,kerning=true,spacing=true,stretch=0, shrink=80]{microtype}

\usepackage{ellipsis}
\usepackage{xspace}
\usepackage{hfoldsty}

\usepackage{ifthen}
\newcommand{\CC}[1][]{$\text{C\hspace{-.25ex}}^{_{_{_{++}}}}
	\ifthenelse{\equal{#1}{}}{}{\text{\hspace{-.625ex}#1}}$}

\makeatletter
\def\thmt@refnamewithcomma #1#2#3,#4,#5\@nil{%
\@xa\def\csname\thmt@envname #1utorefname\endcsname{#3}%
\ifcsname #2refname\endcsname
\csname #2refname\expandafter\endcsname\expandafter{\thmt@envname}{#3}{#4}%
\fi
}
\makeatother
\usepackage[capitalise,noabbrev]{cleveref}   
\usepackage{times}

\newcommand{\new}[1]{\emph{#1}}

\newcommand{\cO}{\ensuremath{{\mathcal O}}\xspace}

\newcommand{\bbR}{\ensuremath{\mathbb{R}}}

\newcommand{\NN}{\mathbb{N}}

\newcommand{\oms}{\{\!\!\{}
\newcommand{\cms}{\}\!\!\}}

\newcommand{\citet}[1]{\citeauthor{#1}~\shortcite{#1}}

\title{The Power of the Weisfeiler-Leman Algorithm for Machine Learning with Graphs}
\author{
Christopher Morris$^1$ \and Matthias Fey$^2$ \And Nils M.~Kriege$^3$
\affiliations
$^1$CERC in Data Science for Real-Time Decision-Making, Polytechnique Montr{\'e}al\\
$^2$Department of Computer Science, TU Dortmund University\\
$^3$University of Vienna, Faculty of Computer Science, Vienna, Austria 
\emails
chris@christopher.morris.info, matthias.fey@udo.edu, nils.kriege@univie.ac.at
}

\begin{document}

\maketitle

\begin{abstract}
In recent years, algorithms and neural architectures based on the Weisfeiler-Leman algorithm, a well-known heuristic for the graph isomorphism problem, emerged as a powerful tool for (supervised) machine learning with graphs and relational data. Here, we give a comprehensive overview of the algorithm's use in a machine learning setting. We discuss the theoretical background, show how to use it for supervised graph- and node classification, discuss recent extensions, and its connection to neural architectures. Moreover, we give an overview of current applications and future directions to stimulate research.
\end{abstract}

\section{Introduction}
Graph-structured data is ubiquitous across application domains ranging from chemo- and bioinformatics~\cite{Barabasi2004,Sto+2020} to computer vision~\cite{Sim+2017}, and social network analysis~\cite{Eas+2010}. To develop successful machine learning models in these domains, we need techniques to exploit the rich information inherent in the graph structure and the feature information within nodes and edges. Due to the highly non-regular structure of real-world graphs, we first have to learn a vectorial representation of each graph or node to apply standard machine learning tools such as support vector machines or neural networks. Here, numerous approaches have been proposed in recent years---most notably, approaches based on \new{graph kernels}~\cite{Kri+2019} or neural architectures~\cite{Gil+2017,Cha+2020}. Especially, graph kernels based on the \new{Weisfeiler-Leman algorithm}~\cite{Wei+1968}, and corresponding neural architectures, known as \new{Graph Neural Networks} (GNNs), have recently advanced the state-of-the-art in supervised node- and graph representation learning.

\begin{figure}[t]
  \begin{center}
    \includegraphics[scale=0.6]{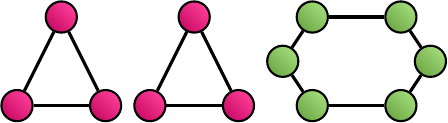}
  \end{center}
  \caption{Two graphs (pink and green) that cannot be distinguished by the $1$-WL.}\label{tria}
\end{figure}

The ($1$-dimensional) Weisfeiler-Leman ($1$-WL) or \new{color refinement} algorithm is a well-known heuristic for deciding whether two graphs are isomorphic: Given an initial \emph{coloring} or \emph{labeling} of the nodes of both graphs, e.g., their degree or application-specific information, in each iteration, two nodes with the same label get different labels if the number of identically labeled neighbors is not equal. If, after some iteration, the number of nodes annotated with a certain label is different in both graphs, the algorithm terminates, and we conclude that the two graphs are not isomorphic. This simple algorithm is already quite powerful in distinguishing non-isomorphic graphs~\cite{Bab+1980}, and has been therefore applied in many areas~\cite{Gro+2014,Ker+2014,Li+2016,Yao+2015,Zha+2017}. On the other hand, it is easy to see that the algorithm cannot distinguish all non-isomorphic graphs~\cite{Cai+1992}. For example, it cannot distinguish graphs with different triangle counts, cf.~\cref{tria}, which is an important feature in social network analysis. Therefore, it has been generalized to $k$-tuples leading to a more powerful graph isomorphism heuristic, which has been investigated in depth by the theoretical computer science community~\cite{Cai+1992,Kie+2016,Bab+2016,Gro2017}. In~\citet{She+2011}, the $1$-WL was first used to obtain a graph kernel, the so-called \new{Weisfeiler-Lehman subtree kernel}. The kernel's idea is to compute the $1$-WL for a fixed number of steps, resulting in a color histogram or feature vector for each graph. The kernel is then computed by taking the pairwise inner product between these vectors. Hence, the kernel measures the similarity between two graphs by counting the number of common colors in all refinement steps. We note here that similar approaches are also used in chemoinformatics~\cite{Rogers2010}.

Graph kernels were dominant in graph classification for several years, which led to new state-of-the-art results on many graph classification tasks. However, they are limited because they cannot effectively adapt their feature representations to a given data distribution since they generally rely on a fixed set of pre-computed features. GNNs have emerged as a machine learning framework addressing the above limitations.
Primarily, they can be viewed as a neural version of the $1$-WL algorithm, where continuous feature vectors replace colors, and neural networks are used to aggregate over local node neighborhoods~\cite{Ham+2017,Gil+2017}. Recently, there has been progress in connecting the two above algorithms. That is, it has been shown that any possible GNN architecture cannot be more powerful than the $1$-WL in terms of distinguishing non-isomorphic graphs~\cite{Mor+2019,Xu+2018b}. Moreover, this line of work has been extended by deriving more powerful neural architectures, e.g., based on the $k$-dimensional Weisfeiler-Leman algorithm, e.g.,~\cite{Mor+2019,Mor+2020} and \cite{Mar+2019}, some of which we survey below.

\paragraph{Present Work.}
Here, we give a comprehensive overview of the applications of the $1$-WL and GNNs in machine learning for graph- and relational data.  We discuss the theoretical background, show how they can be used for supervised graph- and node classification, and discuss recent extensions and its connections. Moreover, we give an overview of current applications and future directions to stimulate research.

\begin{figure*}[t]
  \begin{center}
    \includegraphics[scale=0.86]{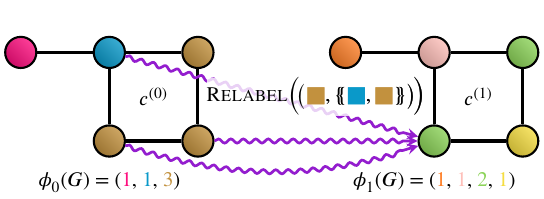}\label{fig:wl}
\hspace{2pt}
    \includegraphics[scale=0.86]{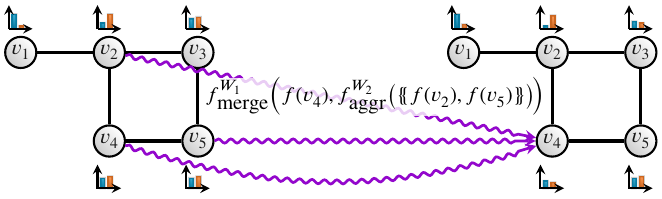}\label{fig:gnn}
  \end{center}
  \caption{Illustration of the update rules of the $1$-WL and Graph Neural Networks.}\label{wli}
\end{figure*}

\section{Background}

A \new{graph} $G$ is a pair $(V,E)$ with a \emph{finite} set of
\new{nodes} $V(G)$ and a set of \new{edges} $E(G) \subseteq \{ \{u,v\}
\subseteq V \mid u \neq v \}$. For ease of notation, we denote an edge $\{u,v\}$ as $(u,v)$. A \new{labeled graph} $G$ is a triple
$(V,E,l)$ with a label function $l \colon V(G) \to \Sigma$,
where $\Sigma$ is some finite alphabet. Then $l(v)$ is a
\new{label} of $v$ for $v$ in $V(G)$. 
The \new{neighborhood} of $v$ in $V(G)$ is denoted by $N(v) = \{ u \in V(G) \mid (v, u) \in E(G) \}$. 
We say that two graphs $G$ and $H$ are \new{isomorphic} if there exists an edge preserving bijection $\varphi \colon V(G) \to V(H)$, i.e., $(u,v)$ is in $E(G)$ if and only if $(\varphi(u),\varphi(v))$ is in $E(H)$ for all $u,v$ in $V(G)$. In the case of labeled graphs, we additionally require that
$l(v) = l(\varphi(v))$ for $v$ in $V(G)$.
We define $[n] = \{ 1, \dotsc, n \} \subset \NN$ for $n \geq 1$, and denote a multiset by $\{\!\!\{ \dots\}\!\!\}$.

A \emph{kernel} on a non-empty set $\mathcal{X}$ is a positive semidefinite function
$k \colon \mathcal{X} \times \mathcal{X} \to \mathbb{R}$.
Equivalently, a function $k$ is a kernel if there is a \emph{feature map} 
$\phi \colon \mathcal{X} \to \mathcal{H}$ to a Hilbert space $\mathcal{H}$ with inner product 
$\langle \cdot, \cdot \rangle$, such that 
$k(x,y) = \langle \phi(x),\phi(y) \rangle$ for all $x$ and $y$ in $\mathcal{X}$.
Let $\mathcal{G}$ be the set of all graphs, then a positive semidefinite function $\mathcal{G} \times \mathcal{G} \to \mathbb{R}$ is a \emph{graph kernel}. 

\subsection{The Weisfeiler-Leman Algorithm}

We now formally describe the {$1$-WL} algorithm for labeled graphs. Let $G = (V,E,l)$ be a labeled graph, in each iteration, $t \geq 0$, the $1$-WL computes a node coloring $c^{(t)}_l \colon V(G) \to \Sigma$,
which depends on the coloring of the neighbors. In iteration $t>0$, we set
\begin{equation*}\label{eq:wlColoring}
c_{l}^{(t)}(v) \!=\! \textsc{Relabel}\Big(\!\big(c_l^{(t-1)}(v),\oms c_l^{(t-1)}(u)\!\mid\!u \in\!N(v) \!\cms \big)\! \Big),
\end{equation*}
where $\textsc{Relabel}$ injectively maps the above pair to a unique value in $\Sigma$, which has not been used in previous iterations. That is, in each iteration the algorithm computes a new color for a node based on the colors of its direct neighbors, cf.~\cref{wli} for an illustration. In iteration $0$, we set $c^{(0)}_l = l$. Hence, after $k$ iterations the color of a node reflects the structure of its
$k$-hop neighborhood. To test if two graphs $G$ and $H$ are non-isomorphic, we run the above algorithm in ``parallel'' on both graphs. If the two graphs have a different number of nodes with a specific color, the \textsc{$1$-WL} concludes that the graphs are not isomorphic. If the number of colors between two iterations does not change, i.e., the cardinalities of the images of $c_l^{(t-1)}$ and $c_l^{(t)}$ are equal, the algorithm terminates. Termination is guaranteed after at most $\max \{ |V(G)|,|V(H)| \}$ iterations~\cite{Gro2017}. 

\paragraph{A Hierarchy of More Powerful Algorithms.}
The $k$-WL was first proposed by L{\'a}szl{\'o} Babai, cf.~\cite{Cai+1992}, based on algorithms proposed by Weisfeiler and Leman, cf.~\cite{Wei+1968,Wei+1976}. To make the algorithm more powerful, it colors tuples from $V(G)^k$ for $k > 1$ instead of nodes. By defining a neighborhood between these tuples, we can define an update rule similar to the $1$-WL. Formally, the algorithm computes a coloring $c^{(t)}_{k,l} \colon V(G)^k \to \Sigma$, and we define the $j$-th neighborhood
\begin{equation*}\label{gnei}
	N_j(s) \!=\! \{ ( s_1, \dotsc, s_{j-1}, r, s_{j+1}, \dotsc, s_k) \mid r \in V(G) \}
\end{equation*}
of a $k$-tuple $s = (s_1, \dotsc, s_k )$ in $V(G)^k$. That is, the $j$-th neighborhood $N_j(s)$ of $s$ is obtained by replacing the $j$-th component of $s$ by every node from $V(G)$. In iteration $0$, the algorithm labels each $k$-tuple with its \new{atomic type}, i.e., two $k$-tuples $s$ and $s'$ in $V(G)^k$ get the same color if the map $s_i \mapsto s'_i$ induces a (labeled) isomorphism between the subgraphs induced by the nodes from $s$ and $s'$, respectively. For iteration $t > 0$, we define 
\begin{equation}\label{labelk}
	C^{(t)}_j(s) = \textsc{Relabel}_{}\big(\oms c^{(t-1)}_{k,l}(s') \mid s' \in N_j(s)\cms\big), 
\end{equation}
and set 
\begin{equation*}
	c_{k,l}^{(t)}(s)\!=\!\textsc{Relabel}\Big( \!\big(c_{k,l}^{(t-1)}(s), \big( C^{(t)}_1(s), \dots, C^{(t)}_k(s)  \big) \! \Big).
\end{equation*}

Hence, two tuples $s$ and $s'$ with $c_{k,l}^{(t-1)}(s) = c_{k,l}^{(t-1)}(s')$ get different colors in iteration $t$ if there exists $j$ in $[k]$ such that the number of $j$-neighbors of $s$ and $s'$, respectively, colored with a certain color is different. The algorithm then proceeds analogously to the $1$-WL.

By increasing $k$, the algorithm gets more powerful in distinguishing non-isomorphic graphs, i.e., for each $k\geq 2$, there are non-isomorphic graphs distinguished by the ($k+1$)-WL but not by the $k$-WL~\cite{Cai+1992}. We note here that the above variant is not equal to the \emph{folklore} variant of $k$-WL described in~\cite{Cai+1992}, which differs slightly in its update rule. It holds that the $k$-WL using~\cref{labelk} is as powerful as the folklore $(k\!-\!1)$-WL~\cite{Gro2021}.

\paragraph{Properties of the Algorithm.} The Weisfeiler-Leman algorithm constitutes one of the earliest approaches to isomorphism testing~\cite{Wei+1976,Wei+1968,Imm+1990} and has been heavily investigated by the theory community over the last few decades~\cite{Gro2017}. Moreover, the fundamental nature of the $k$-WL is evident from a variety of connections to other fields such as logic, optimization, counting complexity, and quantum computing. The power and limitations of $k$-WL can be neatly characterized in terms of logic and descriptive complexity~\cite{Imm+1990}, Sherali-Adams relaxations of the natural integer linear program for the graph isomorphism problem~\cite{Ast+2013,GroheO15,Mal2014}, homomorphism counts~\cite{Del+2018}, and quantum isomorphism games~\cite{Ats+2019}. In their seminal paper,~\citet{Cai+1992} showed that for each $k > 1$ there exists a pair of non-isomorphic graphs of size $\cO(k)$ each that the $k$-WL cannot distinguish and the $(k+1)$-WL can dinstinguish. \citet{Gro2017} gives a thorough overview of these results. For $k=1$, the power of the algorithm has been completely characterized~\cite{Arv+2015,Kie+2015}.  Moreover, upper bounds on the running time~\cite{Ber+2013},  and the number of iterations for $k=1$~\cite{Kie+2020}, and the number of iterations for $k=2$ (folklore variant)~\cite{Kie+2016,Lic+2019} have been shown. For $k=1$ and $2$,~\citet{Arv+2019} studied the abilities of the (folklore) $k$-WL to detect and count fixed subgraphs, extending the work of~\citet{Fue+2017}. The former was refined in~\citet{Che+2020a}. The algorithm (for logarithmic $k$) plays a prominent role in the recent result of \citet{Bab+2016} improving the best-known running time for solving the graph isomorphism problem. Recently,~\citet{Gro+2020a} introduced the framework of Deep Weisfeiler Leman algorithms, which allow the design of a more powerful graph isomorphism test than Weisfeiler-Leman type algorithms.

\section{Applications to Machine Learning}

We now formally define  the \new{Weisfeiler-Lehman subtree kernel}~\cite{She+2011}. The idea is to compute the $1$-WL for $h \geq 0$ iterations resulting in a coloring $c^{(i)}_{l} \colon V(G) \to \Sigma_i,$ for each iteration $i$. After each iteration, we compute a \new{feature vector} $\phi_i(G)$ in $\bbR^{|\Sigma_i|}$ for each graph $G$. Each component $\phi_i(G)_{c}$ counts the number of occurrences of nodes labeled by $c$ in $\Sigma_i$. The overall feature vector $\phi_{\text{WL}}(G)$ is defined as the concatenation of the feature vectors over all $h$ iterations, i.e., $\phi_{\text{WL}}(G) = \big[\phi_0(G), \dots, \phi_h(G) \big].$ The corresponding kernel for $h$ iterations then is computed as  $k_{\text{WL}}(G,H) = \langle \phi_{\text{WL}}(G), \phi_{\text{WL}}(H) \rangle,$ where $\langle \cdot, \cdot \rangle$ denotes the standard inner product.  The running time for a single feature vector computation is in $\cO(hm)$ and $\cO(Nhm+N^2hn)$ for the computation of the Gram matrix for a set of $N$ graphs~\cite{She+2011}, where $n$ and $m$ denote the maximum number of nodes and edges, respectively, over all $N$ graphs. The algorithm scales well to large graphs and datasets, and can be used together with linear SVMs to avoid the quadratic overhead of computing the Gram matrix. For bounded-degree graphs, the kernel can be even approximated in constant time~\cite{Mor+2017}, i.e., the running time is independent of the number of nodes of the graph. 

\paragraph{Variants.} The \new{Weisfeiler-Lehman optimal assignment kernel} is defined as the weight of an optimal assignment between the vertices of two graphs, where the similarity between pairs of vertices is determined by their $1$-WL colors~\cite{Kri+2016}. Similarly, the \new{Wasserstein Weisfeiler-Lehman graph kernel}~\cite{Tog+2019} is obtained from the Wasserstein distance wrt.\@ a ground metric obtained from $1$-WL. \citet{Nik+2017} introduced the concept of optimal assignments in the neighborhood aggregation step. \citet{Rie+2019} combined the  $1$-WL with persistent homology to extract topological features such as cycles. Finally, \citet{Nguyen+2020} leveraged the link between WL features and graph homomorphisms, established by~\citet{Del+2018}, to define graph kernels.

\paragraph{Continuous Information.} Most real-world graphs have attributes, mostly real-valued vectors, associated with their nodes and edges. For example, atoms of chemical molecules have physical and chemical properties, individuals in social networks have demographic information, and words in documents carry semantic meaning. The $1$-WL and the corresponding kernel are discrete. That is, they can only deal with a discrete (countable) label alphabet. Hence, two nodes are regarded as similar if and only if they exactly match, structure-wise and attribute-wise. However, in most applications, it is desirable to compare real-valued attributes with more nuanced similarity measures such as the Gaussian RBF kernel. Hence, extensions of the $1$-WL kernel have been proposed that can deal with continuous information~\cite{Ors+2015,Tog+2019}. For example, a basic instantiation of the \new{GraphInvariant} kernel~\cite{Ors+2015}, can be expressed as
\begin{equation*}
	k_{\text{WV}}(G,H) =\!\!\! \sum_{v \in V(G)} \sum_{v' \in V(H)}\!\! k_V(v,v') \cdot k_W(c_{l}^{(h)}(v),c_{l}^{(h)}(v')).
\end{equation*}
Here, $k_V$ is a user-specified kernel comparing (continuous) node attributes, and $k_W$ is
a kernel determining a weight for a node pair based on their color given by the $1$-WL after $h$ iterations. However, due to the quadratic overhead of computing the above kernel for each pair of graphs, the algorithm does not scale to large datasets. Therefore, \citet{Mor+2016} introduced a scalable framework to compare attributed graphs. The idea is to iteratively turn the continuous attributes of a graph into discrete labels using randomized hash
functions. This allows applying fast explicit graph feature maps, which are limited to graphs with discrete annotations such as the one associated with the Weisfeiler-Lehman subtree kernel. For special hash functions, the authors obtain approximation results for several state-of-the-art kernels handling continuous information. Moreover, they derived a variant of the Weisfeiler-Lehman subtree kernel, which can handle continuous attributes.

\paragraph{Global Information.} Due to the purely local nature of the $1$-WL, it might miss essential patterns in the given data. Hence, recently, graph kernels based on the $k$-WL have been proposed. In~\citet{Mor+2017}, a set-based version of the $k$-WL is employed to derive a kernel. More recently, a local variant of the $k$-WL is proposed~\cite{Morris2020b}, which considers a subset of the original neighborhood in each iteration. The cardinality of this \new{local neighborhood} only depends on the graph's sparsity. The authors showed that the local algorithm has at least the same power as the original $k$-WL, prevents overfitting, and leads to state-of-the-art results on standard benchmark datasets~\cite{Mor+2020}.

\section{Graph Neural Networks}

Intuitively, GNNs compute a vectorial representation, i.e., a $d$-dimensional vector, for each node in a graph by aggregating information from neighboring nodes. Each layer of a GNN aggregates local neighborhood information, i.e., neighbors' features, within each node and then passes this aggregated information on to the next layer. Following~\cite{Gil+2017}, in full generality, a new feature $f^{(t)}(v)$ for a node $v$ is computed as
\begin{equation}\label{eq:gnngeneral}
	f^{W_1}_{\text{merge}}\Big(f^{(t-1)}(v) ,f^{W_2}_{\text{aggr}}\big(\oms f^{(t-1)}(w) \mid  w \in N(v)\cms \big)\!\Big),
\end{equation}
where $f^{W_1}_{\text{aggr}}$ aggregates over the multiset of neighborhood features and $f^{W_2}_{\text{merge}}$ merges the node's representations from step $(t-1)$ with the computed neighborhood features.
Both $f^{W_1}_{\text{aggr}}$ and $f^{W_2}_{\text{merge}}$ may be arbitrary differentiable functions with parameters $W_1$ and $W_2$. See \cref{wli} for an illustration of the architecture. A vector representation $f_{\text{GNN}}$ over the whole graph $G$ can be computed by aggregating the vector representations computed for all nodes, i.e., $f_{\text{GNN}}(G) = \sum_{v \in V(G)} f^{(T)}(v),$
where $T > 0$ denotes the last layer. More refined approaches use differential pooling operators based on sorting~\cite{Zha+2018} or soft assignments~\cite{Yin+2018}. Efficient GPU-based implementations of many GNN architectures can be found in~\cite{Fey+2019}, \cite{Gra+2020}, and~\cite{Wan+2019}.

\paragraph{Connections to the Weisfeiler-Leman Algorithm.}
A recent line of work~\cite{Mor+2019,Xu+2018b,Mar+2019} connects the expressivity of GNNs to that of the $1$-WL algorithm. The results show that GNN architectures do not have more power to distinguish between non-isomorphic (sub-)graphs than the $1$-WL. More formally, let $f^{W_1}_{\text{merge}}$ and $f^{W_2}_{\text{aggr}}$ be any two functions chosen in \cref{eq:gnngeneral}. For every encoding of the labels $l(v)$ as vectors $f^{(0)}(v)$, and for every choice of $W_1$ and $ W_2$, we have that if the GNN parameterized by the above weights distinguishes a pair of graphs, the $1$-WL also will. On the positive side, it has been shown that there exists a GNN architecture and corresponding weights such that it has the same power as the $1$-WL. That is, we get the following insight.

\begin{insight}{\cite{Mor+2019,Xu+2018b}}\label{equal}
	Any possible graph neural network architecture can be at most as powerful as the $1$-WL in terms of distinguishing non-isomorphic graphs.
	
	A GNN architecture has the same power as the $1$-WL if the functions $f^{W_1}_{\text{merge}}$ and $f^{W_2}_{\text{aggr}}$ are \emph{injective}.
\end{insight}
For a detailled discussion, see~\cite{Gro2021}. Hence, in light of the above results, GNNs may be viewed as an extension of the $1$-WL, which has the same power but is more flexible in adapting to the learning task at hand and can handle continuous node features. Moreover, the above results have been generalized to the $k$-WL. That is, \citet{Mar+2019} and \citet{Morris2020b}  derived neural architectures with the same power as the former.
\begin{insight}{\cite{Mar+2019}}\label{equal}
	There exists a GNN architecture that has the same power as the $(k+1)$-WL  in terms of distinguishing non-isomorphic graphs.
\end{insight}
Recently, designing GNNs that overcome the limitations of the $1$-WL received much attention.
\paragraph{Provably Powerful Graph Neural Networks.}
\citet{Mor+2019} extended the expressivity by proposing a higher-order GNN layer, which passes messages between subgraphs instead of vertices by defining a suitable notion of neighborhood between subgraphs. \citet{Mur+2019,Vig+2020} utilize vertex identifiers as node features to maintain information about which vertex in the receptive field has contributed to the aggregated information, leading to provably more powerful architectures. A similar idea was utilized in \citet{Sat+2020}, which includes additional random vertex features instead of vertex identifiers to the message passing phase. The latter was refined by \citet{Das+2020,Abb+2020} which investigated the connection between random coloring and universality. Finally, \citet{botsas2020improving} used higher-order topology as features, while \citet{li2020distance,You+2019} encoded distance information. These generalized GNN architectures show promising results, e.g., on regression tasks of quantum-chemical properties of molecules~\cite{Kli+2020,Mor+2019,Morris2020b}. Further,~\cite{Azi+2020} connect the universality of invariant and equivariant neural networks of graphs to the $k$-WL hierarchy.
\section{Applications}
Graphs define a universal language for describing complex, relational data and hence arise in a wide range of domains across supervised, semi-supervised and unsupervised settings and include tasks such as node- and graph classification, link- and community detection, graph similarity, and graph generation.
Many real-world structures can be naturally modeled as a graph, e.g., physical systems, molecular structures, social networks, and knowledge graphs. However, even when there is no explicit graph structure available, underlying relational information can be synthetically induced to strengthen a model's performance, e.g., for vision or natural language tasks. In the following, we provide a (non-exhaustive) list of applications relying on WL-based graph similarity, either by kernels or GNNs approaches.

\subsubsection{Graph Kernels and $1$-WL}

Since the introduction of the Weisfeiler-Lehman subtree kernel, the $1$-WL has been applied to many application areas, ranging from chem- and bioinformatics~\cite{Sto+2019}, to neuro science~\cite{Veg+2013}, link prediction~\cite{Zha+2017}, dimension reduction of systems of linear equations and linear programs~\cite{Gro+2014}, graph matching~\cite{Kriege+2019}, RDF data~\cite{Vries13}, malware detection~\cite{Nar+2016}, and detection of similar computer programs~\cite{Li+2016}. Experimental studies, e.g.,~\cite{Kri+2019,Mor+2020}, show that the kernel is still a useful and competitive baseline on a wide range of graph classification tasks.

\subsubsection{Graph Neural Networks}
Since the recent advancements in deep learning, research on GNN architectures has been thriving, and they are widely used in tasks with an underlying graph structure, e.g., social network  prediction~\cite{Ham+2017}, traffic prediction~\cite{Yu+2018} and recommender systems~\cite{Yin+2018a}.
They also enable reasoning in knowledge graphs~\cite{Sch+2019}, such as answering queries from a knowledge database~\cite{Ren+2020} or for cross-lingual knowledge alignment~\cite{Xu+2019,Fey+2020}.
Furthermore, relational information also occurs in real-world physical systems where objects are represented as nodes and their relations as edges.
Graph-based learning then enables reasoning about those objects, their relations, and physics effectively~\cite{Kip+2018}.

\paragraph{Chemoinformatics.} In cheminformatics, molecules can be naturally represented as graphs and relational reasoning enables the prediction of chemical properties and biological interaction of proteins~\cite{Duv+2015,Gil+2017,Zit+2018,Kli+2020}. Both, pharmaceutical companies and academia, have an increasing interest in GNNs for computer-aided drug discovery~\cite{Wieder+2020}.

\paragraph{Computer Vision.} Graphs have been also becoming popular in the computer vision domain for learning on scene graphs~\cite{Rap+2017}, image keypoints~\cite{Li+2019,Fey+2020}, superpixels~\cite{Mon+2017}, 3D point clouds~\cite{Qi+2017a} and manifolds~\cite{Fey+2018,Han+2019} where relational learning introduces an important inductive bias into the model.
By incorporating both spatial and semantic information, these models tend to outperform their non-relational counterparts to a large extent.

\paragraph{Generative Models.} Applications like discovering new chemical structures involve the need to synthesize real-world graphs using generative modeling.
Here, graph-based learning is either used to encode the graph into a low-dimensional embedding space~\cite{Simonovsky+2018} or to estimate the quality of generated samples in a GAN setup~\cite{DeCao+2018}.
Generative models can be further distinguished by creating adjacency information at once~\cite{DeCao+2018} or sequentially in an autoregressive fashion~\cite{You+2018,Liao+2019}.

\paragraph{Combinatorial Optimization.} Recently, GNNs have been used aiding to solve $\mathsf{NP}$-hard combinatorial optimization problem. For example, in~\citet{Gas+2019,Li+2019b} they are used to guide tree search algorithms for solving combinatorial optimization to optimality. They have also been used for similar tasks in a $Q$-learning setting~\cite{Kha+2017}. For a thorough overview, see~\cite{Que+2021}.

\section{Discussion and Future Directions}
\begin{figure}
	\centering
	\includegraphics[width=1.00\columnwidth]{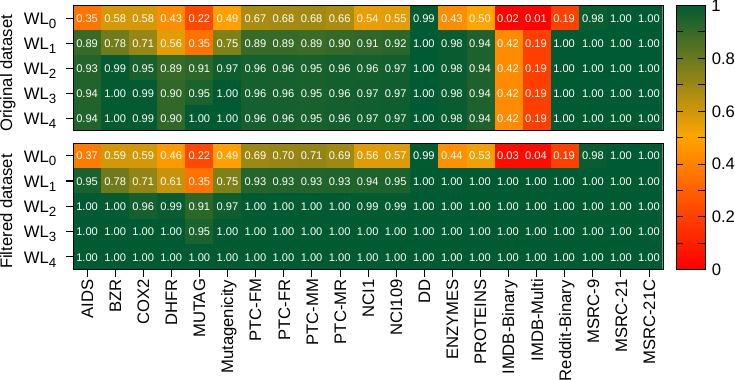}
	\caption{Ability of the 1-WL to distinguish the graphs of common benchmark datasets, before (top) and after (bottom) duplicate filtering.}
	\label{fig:completeness}
\end{figure}
As outline above, Weisfeiler-Leman type methods have been proven to be useful for learning with graphs. However, the learning performance of even the $1$-WL is still poorly understood. To that, we propose several directions to stimulate future research. 

\paragraph{Trading of Expressivity and Generalization.}
Although the $1$-WL's expressivity is well understood, it has not been sufficiently acknowledged that expressivity is not a significant concern for wide-spread benchmark datasets~\cite{Mor+2020}. To confirm this, we have computed the \emph{completeness ratio}, i.e., the fraction of graphs that can be distinguished from all other non-isomorphic graphs in the dataset, see~\ref{fig:completeness}, revealing that the $1$-WL  is sufficiently expressive to distinguish all the non-isomorphic graphs.  Thus, although devising provably powerful graph learning architectures is a meaningful theoretical endeavor, the key to improving real-world tasks is improving GNN's generalization abilities. So far, only a few notable contributions in this direction have been made~\cite{Kri+2018,Gar+2020}. Hence, we believe that more strides should be taken to understand the \emph{generalization performance} of the $1$-WL and GNNs.
\paragraph{Locality.}
The number of iterations of the 1-WL or GNN layers is typically selected by cross-validation and often small, e.g., $\leq5$. For larger values 1-WL's features become too specific, leading to overfitting for graph kernels, while the GNNs' node features become indistinguishable, a phenomenon referred to as \emph{over-smoothing}~\cite{Liu+20}. Moreover, for GNNs, the \emph{bottleneck problem} refers to the observation that large neighborhoods cannot be accurately represented~\cite{Alon2020}.
This prevents both methods from capturing global or long-range information. We believe that the development of new techniques to overcome these issues is a promising research direction.

\paragraph{Incorporating Expert Knowledge.} Nowadays, kernels based on the $1$-WL and GNNs are heavily used in application areas from the life sciences to engineering. However, their usage is often ad-hoc, not leveraging crucial expert knowledge. In cheminformatics, for example, information on functional groups or pharmacophore properties is often available, but it is not straightforward to explicitly incorporate it into the $1$-WL and GNN pipeline. Hence, we view the development of mechanisms to include such knowledge as a crucial step making WL-based learning with graphs more applicable to real-world domains.

\section{Conclusion}

We gave a comprehensive overview of applications of the Weisfeiler-Leman algorithm to machine learning for graph data. Moreover, we introduced GNNs and outlined their connections to the former and applications. We believe that our survey will spark further interest in Weisfeiler-Leman type algorithms, their connections to neural architectures, and their application to solving real-world problems. 

\section*{Acknowledgements}
NMK has been funded by the Vienna Science and Technology Fund (WWTF) through project VRG19-009.
MF has been supported by the \emph{German Research Association (DFG)} within the Collaborative Research Center SFB 876, project A6.

\small
\bibliographystyle{named}
\bibliography{bibliography}

\end{document}